\documentclass[runningheads]{llncs}
\usepackage{graphicx}
\usepackage{amsmath}
\usepackage{amssymb}
\usepackage{bbding}
\usepackage{CJKutf8}
\usepackage{appendix}
\usepackage{multirow}
\usepackage{float}   
\usepackage{url}   
\usepackage[colorlinks=true,linkcolor=blue,citecolor=blue,urlcolor=blue,anchorcolor=blue,]{hyperref}

\begin{document}
\begin{CJK}{UTF8}{gbsn}
\title{Perception Activator: An intuitive and portable framework for brain cognitive exploration}

\titlerunning{Perception Activator}

%
%
\author{Le Xu \and Qi Zhang \and Qixian Zhang  \\ Hongyun Zhang \and Duoqian Miao \and Cairong Zhao} 

\authorrunning{Xu et al.}

\institute{Tongji University}
%
\maketitle              
\begin{abstract}
Recent advances in brain-vision decoding have driven significant progress, reconstructing with high fidelity perceived visual stimuli from neural activity, e.g., functional magnetic resonance imaging (fMRI), in the human visual cortex. Most existing methods decode the brain signal using a two-level strategy, i.e., pixel-level and semantic-level. However, these methods rely heavily on low-level pixel alignment yet lack sufficient and fine-grained semantic alignment, resulting in obvious reconstruction distortions of multiple semantic objects. To better understand the brain's visual perception patterns and how current decoding models process semantic objects, we have developed an experimental framework that uses fMRI representations as intervention conditions. By injecting these representations into multi-scale image features via cross-attention, we compare both downstream performance and intermediate feature changes on object detection and instance segmentation tasks with and without fMRI information. Our results demonstrate that incorporating fMRI signals enhances the accuracy of downstream detection and segmentation, confirming that fMRI contains rich multi-object semantic cues and coarse spatial localization information-elements that current models have yet to fully exploit or integrate.

\end{abstract}
%
%
%

%
%
%
%

\section{Introduction}

How do humans perceive external information? This is a profound and complex question. Although the human brain exhibits substantial anatomical similarity in its functional organization—such as shared properties in memory mechanisms, functional connectivity, and visual cortex function~\cite{FINGELKURTS2005827}—unraveling its intricate workings remains a major challenge. Deciphering perception is not only key to illuminating brain function but also instrumental in developing brain-inspired computational models~\cite{10089190,10.1145/3581783.3611996,9097411}, with wide-ranging applications in clinical diagnostics~\cite{LINDEN2021161} and brain–machine interfaces~\cite{NASELARIS2011400}.
\begin{figure}[!t]
    \centering
    \includegraphics[width=1\linewidth]{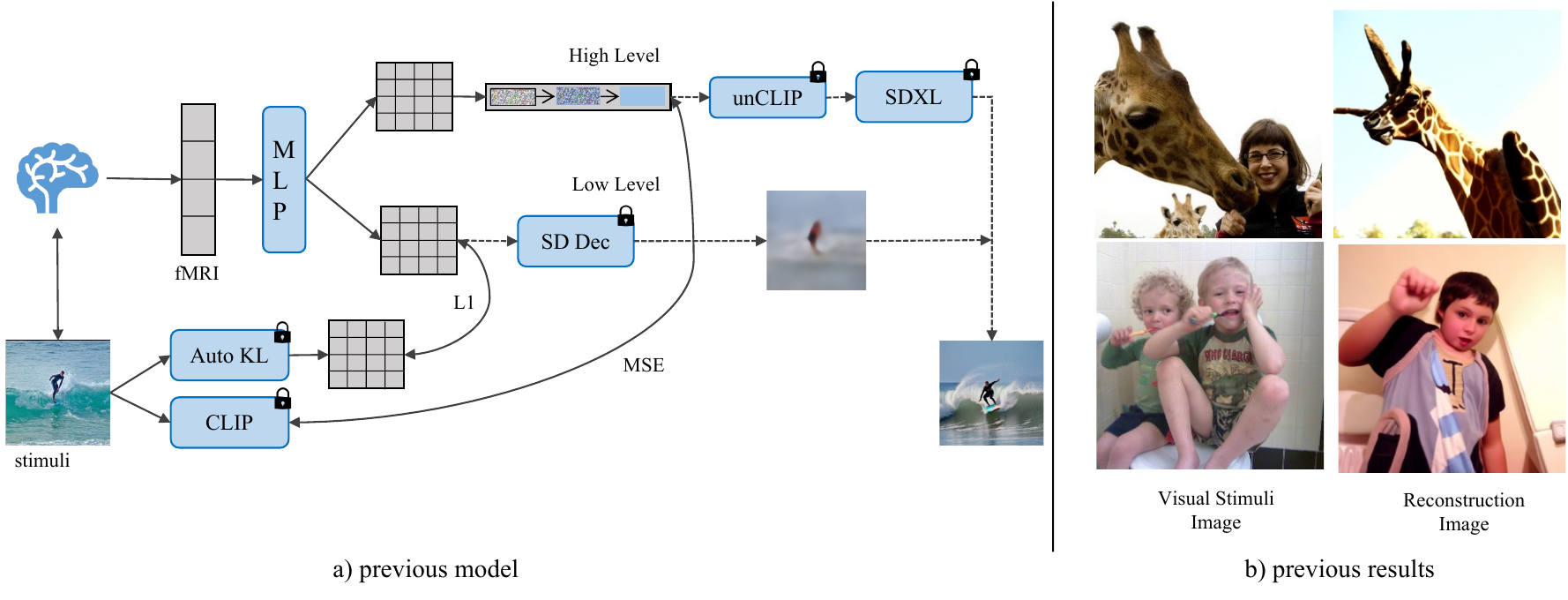}
    \caption{a) The commonly used two-level decoding method. It can be seen that the low-level pixel-level information only plays an auxiliary role in the final step of reconstruction, and the high-level and low-level information are not efficiently fused. b) A comparison between the visual stimulus image and the reconstructed image. It can be seen that the distortion problem of the existing framework occurs on multiple targets。}
    \label{fig1}
\end{figure}
Within this broad research landscape, visual decoding stands out as both pivotal and challenging. It enables us to explore the complex mechanisms underlying visual processing, object recognition, and scene understanding in the brain~\cite{10.5555/3295222.3295390}. Functional magnetic resonance imaging (fMRI) is widely used for noninvasive, spatially precise recording of brain activity to decode natural visual stimuli and reveal cortical representations~\cite{LINDEN2021161}. Open large-scale fMRI datasets—such as the Natural Scenes Dataset (NSD)~\cite{allen2022massive}—have accelerated the adoption of deep learning in visual decoding. Previous studies~\cite{horikawa2016genericdecodingseenimagined,NEURIPS2022_bee5125b,liu2023brainclipbridgingbrainvisuallinguistic,ozcelik2023naturalscenereconstructionfmri,gong2024litemindefficientrobustbrain,bao2024willsalignermultisubjectcollaborative} have tackled classification, retrieval, and reconstruction tasks to extract visual information from fMRI signals, demonstrating the feasibility of reconstructing images that bear resemblance to the original stimuli~\cite{chen2023seeingbrainconditionaldiffusion,NEURIPS2022_bee5125b,10.1145/3581783.3613832,ozcelik2023naturalscenereconstructionfmri,scotti2023reconstructing,10205187}.

As illustrated in Fig.~\ref{fig1}, most state-of-the-art methods employ a two-level decoding strategy: at the high level, mimicking the brain’s perception of coarse semantic information (e.g., category, action), and at the low level, restoring fine-grained pixel-level details (e.g., color, texture, contours). The success of these approaches largely depends on aligning fMRI signals with image representation spaces through pretrained cross-modal models like CLIP~\cite{pmlr-v139-radford21a,scotti2023reconstructing}, Stable Diffusion~\cite{9878449}, and VAEs~\cite{guo2023neuroclipneuromorphicdataunderstanding}. However, when stimuli contain multiple objects, existing models often produce distortions: different object classes may be decoded as the same category, and low-level pixel details are insufficient to correct these errors. In cases with multiple instances of the same category, a single semantic label fails to guide pixel-level reconstruction. This overreliance on high-level semantics, with low-level details playing only a supporting role, leads to “hierarchical disconnection” and reconstruction distortion.

To address this, we construct a cross-attention framework that uses fMRI representations as conditional inputs, injecting voxel signals into multi-scale image features. We validate our approach on two instance-level tasks—object detection and instance segmentation—and show that fusing fMRI signals in the NSD yields a 7 \% improvement in detection accuracy and a 4 \% improvement in segmentation accuracy. These results confirm that fMRI encodes rich multi-object semantic cues and coarse spatial localization information, which, however, existing models have underutilized.

Our main contributions can be summarized as follows:\\
- Conditional Brain-Decoding Framework: Introducing fMRI voxel signals as control conditions via cross-attention injections, providing a systematic tool to analyze their impact on various downstream metrics and representations for probing visual perception mechanisms.\\
- Instance-Level Task Validation: Integrating object detection and instance segmentation tasks into our framework, empirically demonstrating that fMRI signals contain rich instance-level category and spatial details.\\
- LoRA-Based Fine-Tuning: Employing LoRA as a lightweight cross-modal adaptation module to efficiently fine-tune downstream tasks without altering pretrained fMRI representations.\\

\section{Related Work}
\subsection{Brain Visual Decoding}
The use of functional magnetic resonance imaging (fMRI) for visual decoding in the human brain has been a long-standing pursuit, driven by fMRI’s high spatial resolution~\cite{LINDEN2021161}. Early efforts tackled the modality’s low signal-to-noise ratio by employing purely linear methods—such as ridge regression—to embed~\cite{scotti2023reconstructing} voxel responses into a shared intermediate space for coarse-grained image retrieval and reconstruction. Researchers then enriched this pipeline by leveraging hierarchical features extracted from pretrained VGG networks. With the advent of cross-modal pretraining, CLIP ViT emerged as a favored image representation, and methods began aligning fMRI patterns to CLIP’s CLS embeddings or deeper hidden-layer features~\cite{pmlr-v139-radford21a}. Contrastive learning frameworks , variational autoencoders, and diffusion priors were introduced to boost performance. Self-supervised strategies like masked pretraining on BOLD5000 and large MLP backbones augmented with MixCo~\cite{Kim2020MixCoMC} data augmentation and diffusion initialization further advanced high-fidelity reconstruction.

On the architectural and optimization front, the field has diversified from linear regressors and ResNet-enhanced encoders to large-scale MLPs, Transformer-based models, and frequency-domain learners aimed at parameter efficiency and faster inference. Supervision objectives have similarly evolved—from cross-entropy classification and retrieval losses to CLIP alignment, BiMixCo/Soft-CLIP~\cite{scotti2023reconstructing} contrastive schemes, and masked modeling. Moreover, novel pipelines have explored projecting reconstructed image features into text-generation~\cite{gong2024mindtunercrosssubjectvisualdecoding} or image-captioning frameworks, realizing a closed-loop brain-to-language decoding. However, the success of these methods largely depends on aligning the fMRI signals with the representation space of the images, ignoring the role of the fMRI signals themselves, and not exploring the visual information contained in the fMRI signals from the perspective of intuitive understanding.
\begin{figure}[!t]
    \centering
    \includegraphics[width=1\linewidth]{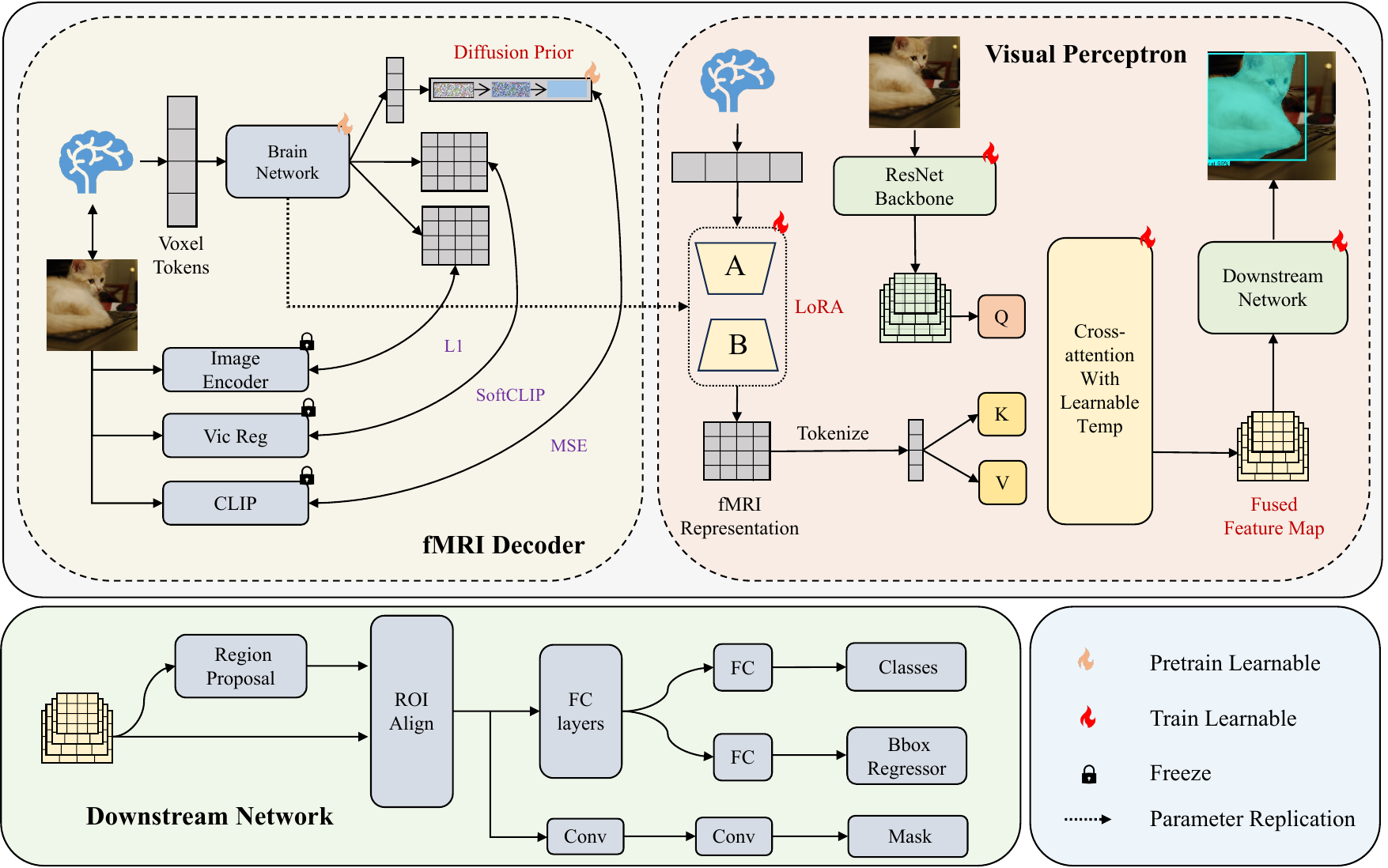}
    \caption{Overview of our framework, consisting of two main components: fMRI Decoder and Visual Perceptron. The Downstream  Network can be adapted to different downstream tasks.}
    \label{fig2}
\end{figure}
\subsection{Instance-level Tasks}
In the research of computer vision, Object Detection and instance segmentation respectively undertake roles at different levels: Object detection aims to identify the categories and approximate positions of each object in the image, corresponding to the coarse-grained perception of "what" and "where" by the brain's visual system; Instance Segmentation further generates pixel-level masks for each target to capture its fine contour features, which is equivalent to fine-grained analysis of "shapes" and "boundaries". Mask R-CNN, as a classic two-stage framework in this field, first locates candidate boxes through the Region Proposal Network (RPN), and then performs classification, border regression and pixel-level mask prediction in parallel, enabling the detection and segmentation tasks to be collaboratively optimized under the same structure.

\section{Method}
As shown in Fig.~\ref {fig2}, our model consists of two parts: 1) \textbf{fMRI Decoder} acquires the high-level and low-level representations of fMRI and fine-tunes them using LoRA. 2) \textbf{Visual Perceptron (VP)} fuses fMRI representations and image features to perform target detection and instance segmentation tasks.

\subsection{fMRI Decoder Pre-training }

\textbf{MLP Backbone.} To achieve high-fidelity pixel-level reconstruction, we align fMRI representations with CLIP’s rich feature space. Specifically, we employ OpenCLIP ViT-bigG/14, whose image token embeddings are of size 256$\times$1664. The preprocessed fMRI inputs, denoted $M$, first pass through a tokenization layer and then through an MLP backbone— comprising four residual blocks— which maps the original $d_{0}$-dimensional input into the 256$\times$1664 embedding space. Formally:
\begin{equation}
Z = \epsilon\bigl(M\bigr).
\end{equation}
where $\epsilon(\cdot)$ represents the MLP backbone mapping. The resulting intermediate representation Z is subsequently fed into three submodules dedicated to retrieval, high-level reconstruction, and low-level reconstruction, respectively.

\textbf{Retrieval Submodules.} To perform the retrieval task, a simple way is to conduct a shallow mapping of the backbone embeddings and supervision with fMRI-to-image CLIP contrastive loss. For scarce fMRI samples, appropriate data augmentation helps model convergence. A recently effective voxel mixture paradigm based on MixCo~\cite{Kim2020MixCoMC} works well. Two raw fMRI voxels $V_{i}$ and $V_{j}$ are mixed into $V_{mix_{i,j}}$ using a factor  sampled from the Beta distribution:
\begin{equation}
\begin{aligned}
V_{\mathrm{mix},i,j}
&= \lambda_i V_i + (1-\lambda_i)\,V_j,\\
M_{\mathrm{mix},i,j}
&= \mathrm{Ridge}^{(s)}\bigl(V_{\mathrm{mix},i,j}\bigr),\\
Z_{\mathrm{mix},i,j}
&= \epsilon\bigl(M_{\mathrm{mix},i,j}\bigr).
\end{aligned}
\end{equation}

where $j$  denotes an arbitrary mixing index in the batch. The forward mixed contrastive loss MixCo is formulated as:
\begin{align}
\mathcal{L}_{\mathrm{MixCo}}
=-\frac{1}{\lvert B\rvert}
\sum_{i=1}^{\lvert B\rvert}
\Biggl[
\lambda_i
\log
\frac{
  \exp\bigl(Z_{\mathrm{mix},i,j}\cdot g_i/\tau\bigr)
}{
  \sum_{m=1}^{\lvert B\rvert}\exp\bigl(Z_{\mathrm{mix},i,j}\cdot g_m/\tau\bigr)
}\\
+
(1-\lambda_i)
\log
\frac{
  \exp\bigl(Z_{\mathrm{mix},i,j}\cdot g_j/\tau\bigr)
}{
  \sum_{m=1}^{\lvert B\rvert}\exp\bigl(Z_{\mathrm{mix},i,j}\cdot g_m/\tau\bigr)
}
\Biggr].
\end{align}
where $g$ denotes the CLIP image embeddings, $\tau$ denotes a temperature hyperparameter, and $B$  is the batch size. Here we use the bidirectional loss $L_{BiMixCo}$.

\textbf{Two-Level Submodule.} Low-level pipeline is widely used to enhance low-level visual metrics in reconstruction images, which map voxels to the latent space of Stable Diffusion’s variational autoencoder(VAE) and is a ’guess’ for the reconstruction. It consists of an MLP and a CNN upsampler with L1 loss in Stable Diffusion’s latent space $z$.

\begin{equation}\mathcal{L}_{lowlevel}=\frac{1}{|B|}\sum_{i=1}^{|B|}|z_{i}-\hat{z_{i}}|.\end{equation}

On the other hand, high-level pipeline emphasizes pixel-wise align ment. Inspired by DALLE·2~\cite{ramesh2022hierarchicaltextconditionalimagegeneration}, a diffusion prior is recognized as an effective means of transforming backbone embeddings into CLIP ViT image embeddings, in which mean square error loss is used:

\begin{equation}\mathcal{L}_{prior}=\frac{1}{|B|}\sum_{i=1}^{|B|}||g_i-\hat{g}_i||_2^2.\end{equation}

Thus, the end-to-end loss for fMRI decoder pre-training is:
\begin{equation}\mathcal{L}_{fMRI}=\mathcal{L}_{prior}+\alpha_{1}\mathcal{L}_{lowlevel}+\alpha_{2}\mathcal{L}_{BiMixCo}.\end{equation}

\subsection{Visual Perceptron}
In this section, we will demonstrate how to integrate brain visual signals (fMRI) into image features and conduct training for downstream tasks. To explore how the brain's visual perception system responds to the targets in visual stimuli, we selected the object detection and instance segmentation tasks in the computer vision task as downstream tasks.

\textbf{Low-Rank Decomposition.}The Low-Rank Adapter (LoRA)~\cite{hu2021loralowrankadaptationlarge} is a method for parameter efficient fine-tuning. It can reduce memory requirements by using a small set of trainable parameters, while not updating the original model parameters which remain fixed. During fine-tuning, the gradients generated by the optimizer are passed through the fixed model to the LoRA. Subsequently, the LoRA can update its own parameters to fit the optimization objective. We denote the decomposition as: 
\begin{equation}\Delta W=BA.\end{equation}
where :$A\in{\mathcal{R}^{r\text{×}d}}$ and $B\in{\mathcal{R}^{d\text{×}r}}$ are two low-rank matrices, with a rank of $r$. Moreover, random initialization is implemented for $A$, and zero initialization is implemented for $B$, enabling the initial output to be zero. Through this module, the modification of the parameters of the original fMRI decoder backbone is avoided during the multimodal fusion process, and its ability to handle downstream tasks itself is not compromised.

\textbf{Cross-Attention Fusion.} To inject fMRI information before RoIAlign at every FPN level, we insert an identical cross-modal attention block on each FPN output feature map. Denote the $l$-th FPN feature as $F^{(l)}\in\mathbb{R}^{N_l\times d}$, where $N_{l}=H_{l}\text{x}W_{l}$  is the number of spatial tokens at that level and d=256 is the channel dimension. Let the fMRI backbone embedding be $Z\in\mathbb{R}^{M\times d}$, where $M=256\text{x}1664$ is the number of fMRI tokens. The cross-attention computation proceeds as follows:

1. Linear projections:
\begin{equation}Q^{(l)}=F^{(l)}W_Q,\quad K=ZW_K,\quad V=ZW_V,\end{equation}
with learnable matrices $W_Q,W_K,W_V\in\mathbb{R}^{d\times d_k}\mathrm{~and~}d_k=d$.

2. Scaled dot-product attention with learnable temperature:

Introduce a learnable scalar temperature $\tau>0$. The attention weights are computed as:
\begin{equation}A^{(l)}=\mathrm{Softmax}\left(\frac{Q^{(l)}K^\top}{\tau\sqrt{d_k}}\right)\quad\in\mathbb{R}^{N_l\times M}.\end{equation}

3. Cross-modal feature aggregation:
\begin{equation}O^{(l)}=A^{(l)}V\quad\in\mathbb{R}^{N_l\times d_k}.\end{equation}

4. Feature fusion with residual connection:
\begin{equation}\widetilde{F}^{(l)}=\mathrm{Layer}\mathrm{Norm}(F^{(l)}+O^{(l)}W_O),\end{equation}
where  $W_O\in{\mathcal{R}^{d_k\text{×}d}}$ projects the aggregated features back to the original dimension.The fused feature $\widetilde{F}^{(l)}$ is then passed to RoIAlign and subsequently to the detection and mask heads. 
Here, we follow the standard Mask R-CNN multi-task loss formulation, summing classification, box-regression, and mask losses over a set of $R$ proposals:
\begin{equation}\mathcal{L}=\mathcal{L}_{\mathrm{det}}+\mathcal{L}_{\mathrm{mask}}\end{equation}

By sharing this cross-attention block across all FPN scales and using a learnable temperature to adapt attention intensity, the model dynamically incorporates brain activity cues into multi-scale visual features, allowing us to assess the contribution of fMRI signals to both detection and instance segmentation.

\section{Experiments}
\subsection{Datasets}
The Natural Scenes Dataset (NSD) is a massive 7T neuroscience dataset encompassing fMRI data from 8 subjects. Throughout the NSD experiment, participants were presented with images sourced from MSCOCO, while their neural responses were recorded utilizing a high-resolution 7-Tesla fMRI scanner. Our study concentrates on Subj01, as this subject completed all 40 session scans. Within the dataset, each subject’s train set comprises 8859 distinct visual stimuli and 24980 fMRI, wherein each image was repetitively presented for viewing 1-3 times. The test set contains 982 visual stimuli and 2770 fMRI. It is crucial to emphasize that within the training set for each subject, the visual stimuli employed do not overlap with those utilized for other subjects. However, the same visual stimuli are employed across subjects within the test set. Upon selecting the cortical area named nsdgeneral, we acquired voxel sequences for subject 1, whose length is 15724. These regions span visual areas ranging from the early visual cortex to higher visual areas. Our experimental setup is consistent with the NSD image reconstruction and retrieval articles~\cite{NEURIPS2022_bee5125b,mai2023unibrainunifyimagereconstruction,ozcelik2023naturalscenereconstructionfmri,scotti2023reconstructing,10205187}.

\subsection{Implementation details}
All of our experiments were conducted on 8 NVIDIA 4090 GPUS with 16GB of video memory. During the pre-training stage of the fMRI decoder, we conducted 150 epochs of training on the data of 40 sessions of subject 1. We use AdamW for optimization, with a learning rate set to 3e-4, to which the OneCircle learning rate schedule was set. During the training process, we use data augmentation from images and blurry images and replace the BiMixCo with SoftCLIP~\cite{scotti2023reconstructing} loss in one-third of the training phase. During the training stage of the visual perceptron, we trained all the visual stimulus images watched by subject 1 for 20 epochs. We use the SGD for optimization, with a learning rate set to 4e-3, to which a StepLR learning rate schedule was set. 

\section{Results and Analysis}
\subsection{Object Detection and Instance Segmentation}
In computer vision, object detection focuses on identifying the class and approximate location of each object in an image, while instance segmentation further generates pixel-level masks to capture fine contour details. This mirrors the human visual system’s coarse-grained perception of “what and where” and its fine-grained parsing of “boundary and shape.” In our study, we compare key performance metrics on these two tasks before and after incorporating fMRI signals, in order to probe how brain activity contributes differently to coarse semantic localization versus fine contour representation.

\textbf{Metrics.} In this experiment, we adopt the COCO benchmark’s unified evaluation metrics—AP (Average Precision), computed as the mean precision at IoU thresholds from 0.50 to 0.95 in steps of 0.05, along with$AP_{50}$, $AP_{75}$, $AP_S/ AP_M/ $ $AP_L$ (for small/medium/large objects), and AR (average recall over the top 100 predictions). IoU is calculated over bounding boxes for detection and over masks for segmentation, thus jointly measuring the model’s performance in localization and pixel-level segmentation.

\begin{table}[]
\caption{Comparison with the baseline methods on NSD dataset.  The bold entities denote the best performance. 
\label{tab:table1}}
\centering
\renewcommand\arraystretch{1.0}
\begin{tabular}{cccccccccccccc}
\hline
\multicolumn{2}{c}{Method}                  & $AP$   & $AP_{50}$ & $AP_{75}$ & $AP_S$ & $AP_M$ & $AP_L$ & $AR_1$ & $AR_{10}$ & $AR_{100}$ & $AR_S$ & $AR_M$ & $AR_L$ \\ \hline
\multirow{2}{*}{Det}  & ResNet50  & 20.6    & 35.5    & \textbf{23.1}    & 5.6   & 28.9   & 47.5  & \textbf{22.0}  & 28.8    & 29.2    & 10.2   & 38.5   & 61.5   \\
                     & our                  & \textbf{20.9}    & \textbf{36.7}    & 20.9    & \textbf{5.8}   & \textbf{29.2}   & \textbf{49.5}   & 21.3  & \textbf{29.2}    & \textbf{30.1}    & \textbf{11.5}   & \textbf{40.3}   & \textbf{61.9}   \\ \hline
\multirow{2}{*}{Seg} & ResNet50  & \textbf{19.3}   & 32.5   & \textbf{20.5}    & \textbf{4.3}   & \textbf{28.4}   & 48.6   & \textbf{20.2}  & 26.0    & 26.2    & 8.7   & 35.6   & 54.8   \\
                     & 0ur                  & 18.8   & \textbf{32.6}    & 20.2    & 4.1   & 27.5   & \textbf{49.0}   & 19.8  & \textbf{26.3}    & \textbf{26.9}    & \textbf{9.2}   & \textbf{36.9}   & \textbf{55.6}   \\ \hline
\end{tabular}
\end{table}

\begin{figure}[!t]
    \centering
    \includegraphics[width=1\linewidth]{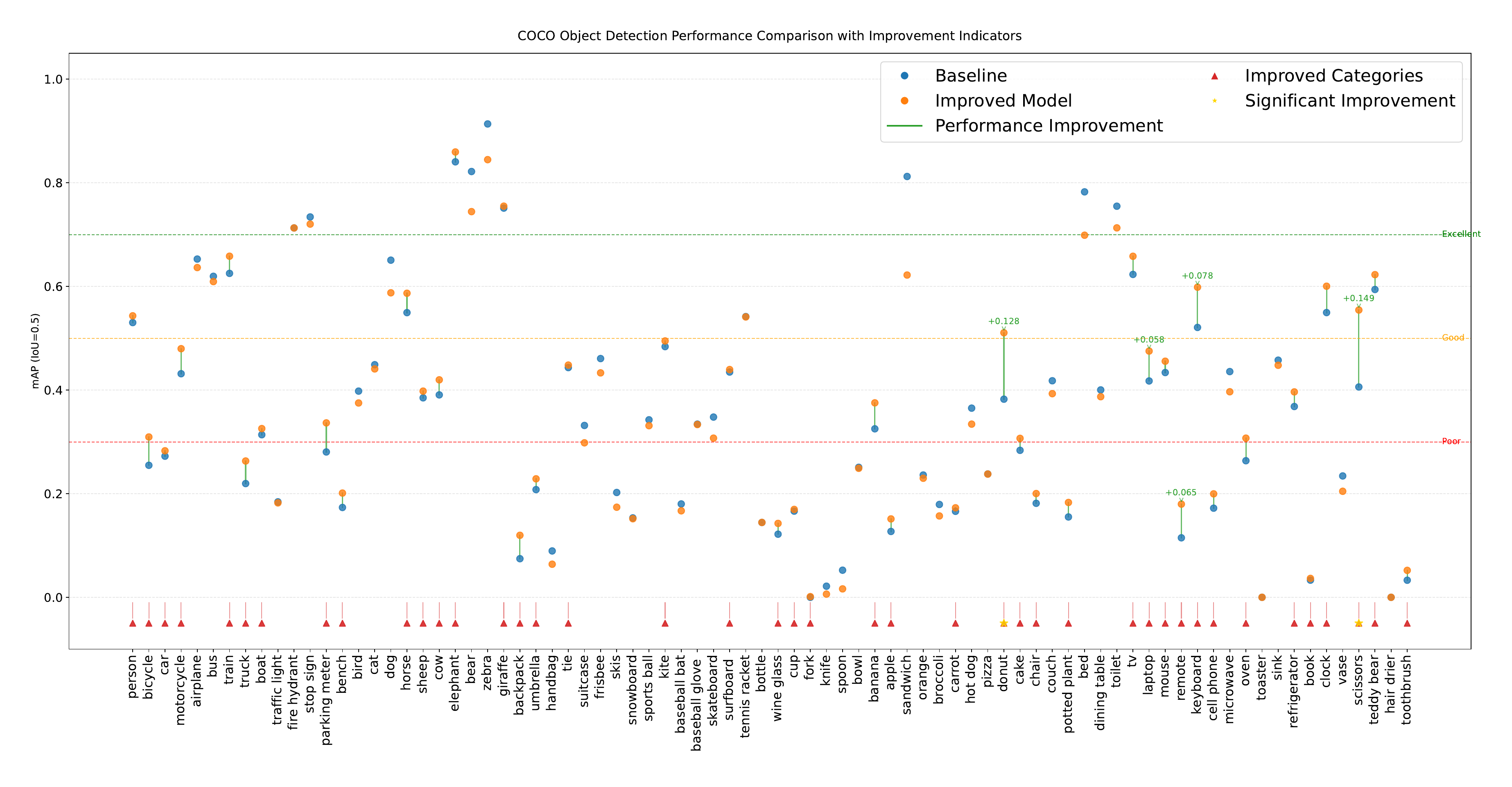}
    \caption{mAP point map presented by category (IoU$>$50). Our model has been improved in most categories. }
    \label{fig3}
\end{figure}

We present both a comprehensive metrics Table~\ref{tab:table1}  and a per-category visualization Fig.~\ref{fig3}. After incorporating fMRI signals, the model shows consistent gains across most detection metrics—most notably, APL increases by 2.0\% and ARₘ by 1.8\% for medium and large objects—indicating that brain activity provides stronger cues for larger targets. Overall, the improvement in AR exceeds that in AP, suggesting that the injected fMRI information chiefly contributes high-level semantic and coarse spatial signals: it helps the model “find” more objects but has limited impact on box precision and confidence, so AP only rises modestly. In instance segmentation, only AR sees a slight boost, implying that fMRI adds little to the fine-grained contour delineation required for accurate masks; it enhances recall via semantic cues but does not improve pixel-level accuracy. In sum, our results indicate that fMRI signals predominantly carry global semantic information. The per-category mAP chart shows that integrating fMRI information yields gains across most classes. Additionally, the fMRI signals appear to enhance the model’s understanding of semantically related categories—such as small objects like scissors or remote controls, which are often held by a human hand—thereby helping the model detect small targets that were previously difficult to recognize.

\subsection{Ablations}
In this section, we perform ablation experiments to explore the performance sources of our study.\\
\textbf{Ablation for components.} We conducted ablation studies on the two pivotal components of our model: the LoRA-finetuned fMRI decoder pretraining strategy and the cross-modal fusion cross-attention module. All experiments were performed under identical hyperparameter settings, with quantitative results summarized in Table~\ref{tab:table2}. The findings demonstrate that both components contribute positively to overall performance: the pretrained fMRI decoder injects rich semantic information during finetuning, thereby enhancing the visual feature representation, while the cross-attention module outperforms naive feature concatenation by avoiding the disruptive effects of directly merging fMRI signals with Mask R-CNN’s native representations. Together, these modules complement each other at different levels, jointly driving improvements in both detection and segmentation tasks.
\begin{table}[]
\caption{Ablation study of fMRI pre-train strategy and cross-attention module. After removing the cross-attention, we use simple feature stitching. The bold entities denote the best performance. 
\label{tab:table2}}
\centering
\tabcolsep=0.4cm
\renewcommand\arraystretch{1.0}
\begin{tabular}{cccccccccccccc}
\hline
\multicolumn{2}{c}{Method}                  & $AP$   & $AP_{50}$ & $AR_1$ & $AR_{100}$  \\ \hline
\multirow{3}{*}{Det}  & w/o fMRI pre-train            & 20.4    & 35.3    & 20.9   & 28.3     \\
                      & w/o cross-attention            & 14.6    & 28.4    & 16.6    & 21.9      \\
                      & our                  & \textbf{20.9}    & \textbf{36.7}    &\textbf{21.3}    & \textbf{29.2}    \\ \hline
\multirow{3}{*}{Seg} & w/o fMRI pre-train             & 18.5   & 32.0   & 19.7    & 25.1     \\
                     & w/o cross-attention            & 13.6    & 25.0    & 15.5    & 20.0      \\
                     & 0ur                  & \textbf{18.8}   & \textbf{32.6}    & \textbf{19.8}   & \textbf{26.9}     \\ \hline
\end{tabular}
\end{table}

\begin{figure}[!b]
    \centering
    \includegraphics[width=1\linewidth]{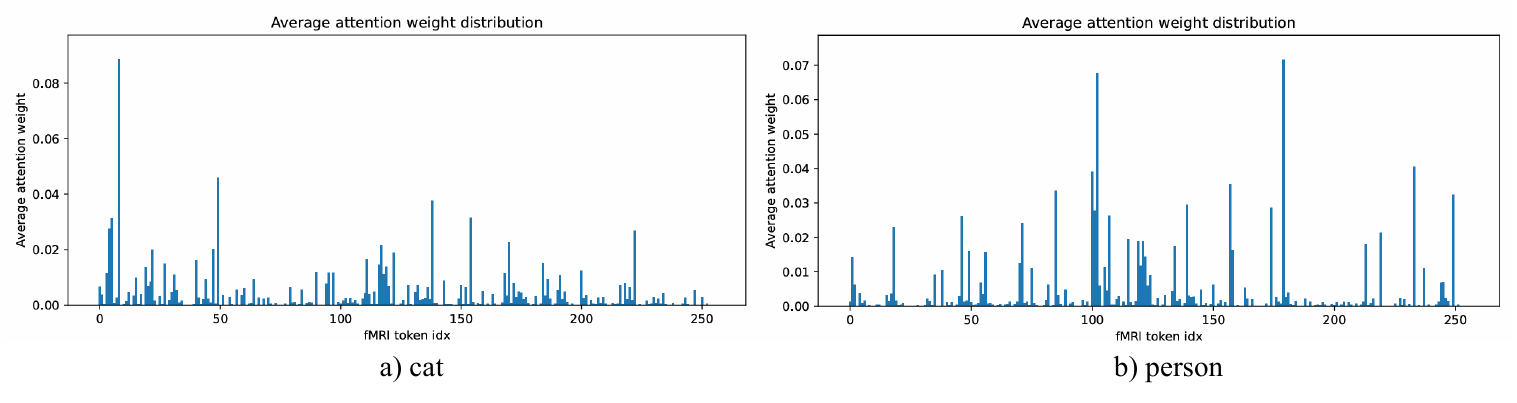}
    \caption{Multi-layer average attention weight map. Here, the visual stimulus image with the main categories of cat and person is selected. }
    \label{fig5}
\end{figure}
\begin{figure}[!t]
    \centering
    \includegraphics[width=0.9\linewidth]{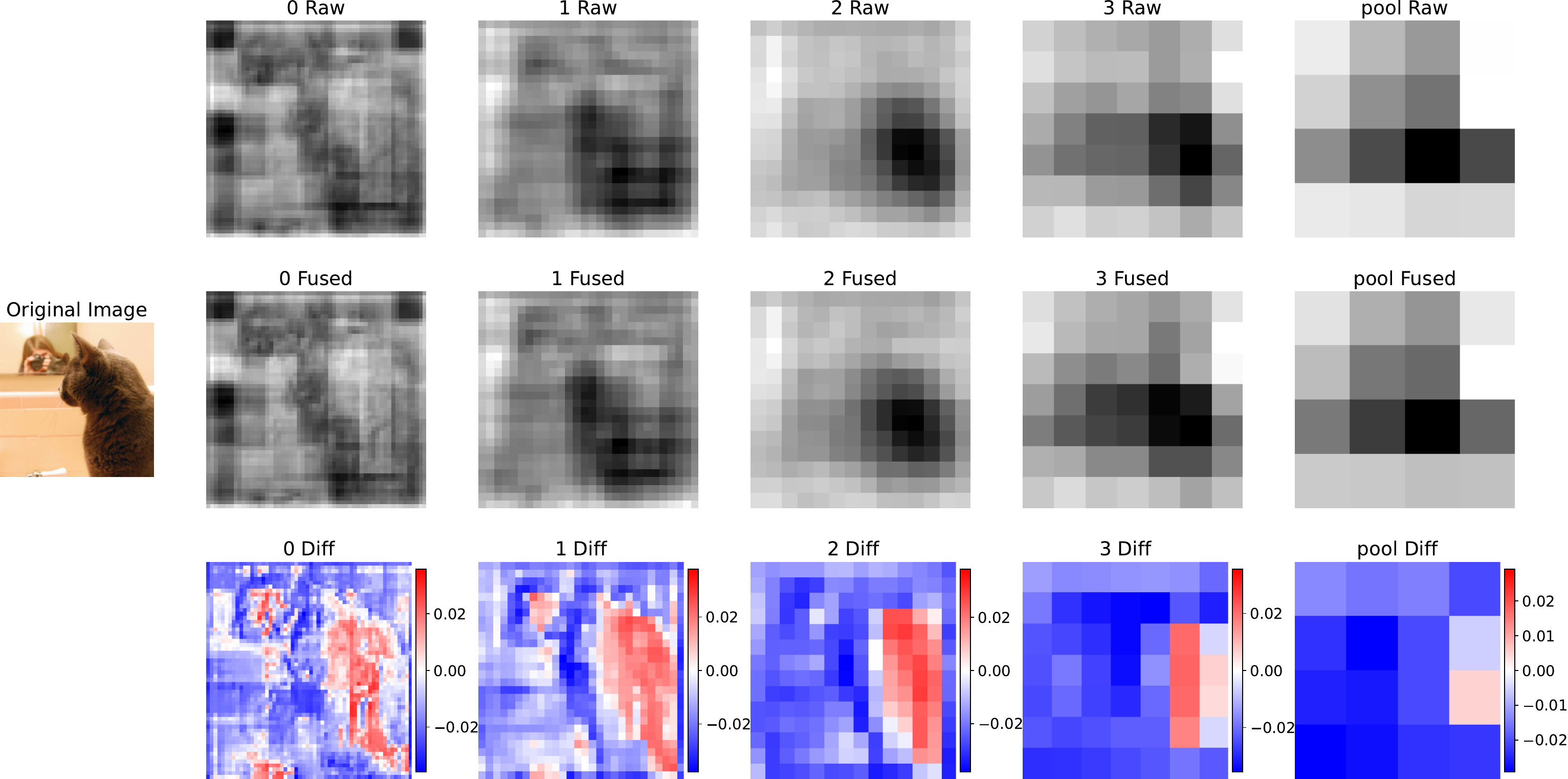}
    \caption{Visualization of the differences before and after feature map fusion and fMRI characterization."Raw" represents the unfused feature map, "fused" represents the fused one, and "diff" is the difference after fusion minus that before fusion, representing the number of layers of the current feature map}
    \label{fig6}
\end{figure}
\subsection{Perception Experiment and Visualization}
In this section, we controlled the category and number of objects in the visual stimuli to examine the sequence of fMRI tokens receiving attention and the regions of original voxels being activated, thereby probing the brain’s visual perception patterns for different targets.\\
\textbf{Cross-Attention Heatmaps.} To investigate the cross-modal fusion mechanism, we visualized how image features attend to fMRI representations within our cross-attention module. Specifically, for each FPN level we computed the attention matrix between image features and fMRI embeddings, then averaged it over the spatial dimensions of the image to obtain, for each fMRI token (or brain region), an overall attention score. The resulting heatmaps reveal, at multiple semantic scales, which brain areas the model prioritizes during fusion, reflecting its dynamic engagement with fMRI signals.

We further conducted a category-differentiation experiment by selecting 10 images each of two object classes—humans and cats—based on the dominant subject in each image. By averaging their attention scores (see Fig.~\ref{fig5}), we observed that images of the same class elicit similar activation patterns across fMRI tokens, while different classes produce distinctly different patterns. This indicates that the fMRI representations contain clear category-specific semantic information and exhibit consistent response signatures.\\
\textbf{Feature Fusion Difference Maps.} Additionally, we visualized the per-layer feature differences before and after cross-modal fusion. For each FPN level, we computed the absolute channel-wise difference and then averaged over the channel dimension to produce a 2D heatmap. As shown in Fig.~\ref{fig6}, the most pronounced differences concentrate on the centers and semantically salient regions of objects, particularly for medium and large targets. In the early, shallow layers, small non-primary objects also exhibit enhanced feature responses, whereas in the deeper layers these signals are suppressed. This indicates that the injected fMRI information not only selectively amplifies high-level features in relevant areas but also carries rich multi-object cues. This visualization corroborates our quantitative findings: cross-modal fusion does not indiscriminately perturb the feature space but instead sharpens the model’s focus on true target locations and semantics, reflecting the rich category and spatial information inherent in fMRI signals.
\begin{figure}[!t]
    \centering
    \includegraphics[width=0.8\linewidth]{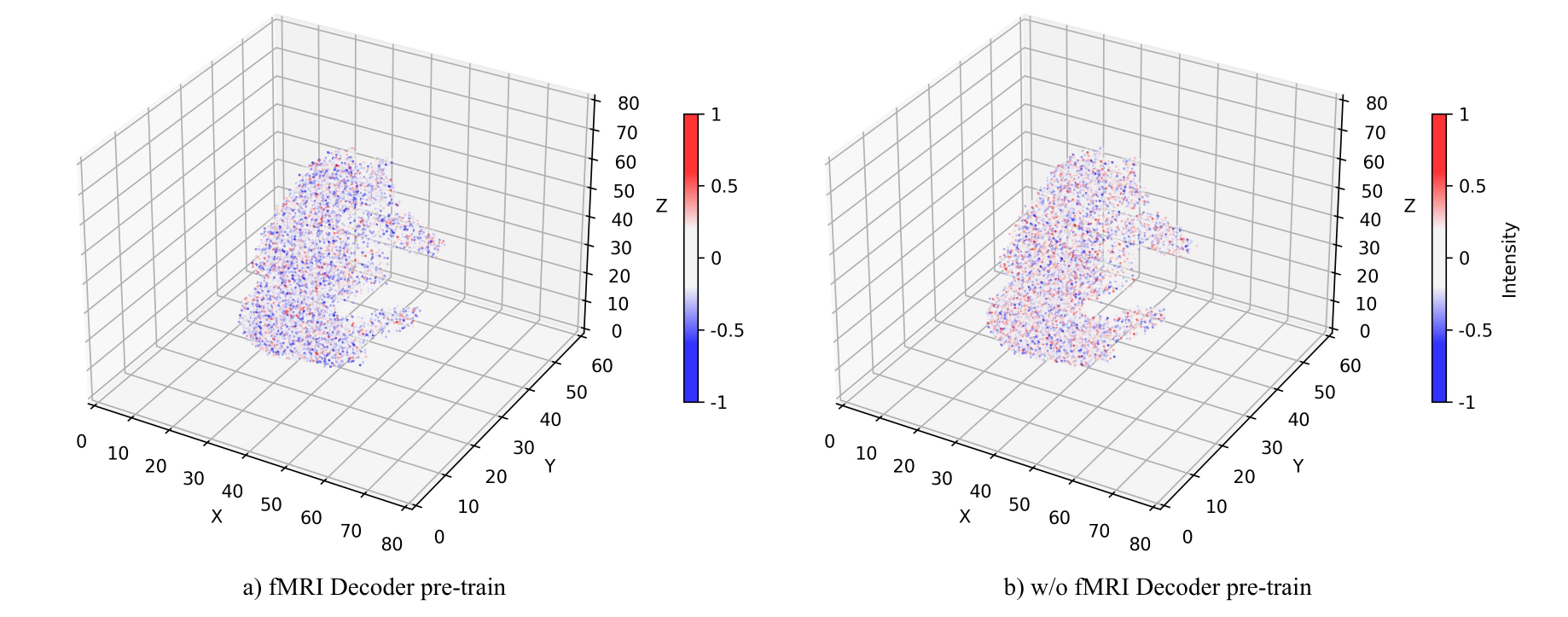}
    \caption{Visualization of the initial layer weights of fMRI. a) is the weight after pre-training, and b) is the weight without pre-training.}
    \label{fig8}
\end{figure}

\textbf{Visualization of the initial layer weights of fMRI} Additionally, we visualized the weight distributions of the initial layer of the fMRI decoder before and after pretraining, as shown in Fig.~\ref{fig8}. The comparison shows that the pretrained model exhibits a sparser activation pattern in its first layer, with most channels suppressed, whereas the no pre-trained model activates many more channels at this stage. Nevertheless, the pretrained model achieves superior downstream performance. We conjecture that pretraining injects precise semantic information into the fMRI representations, allowing the model to rely on fewer activations and thus mitigate noise inherent in the raw fMRI signals. In contrast, the no pre-trained model depends solely on downstream supervision; since instance segmentation is a fine-grained task, it is prone to overfitting on noisy fMRI inputs, resulting in degraded accuracy.

\section{Limitation and Future Work}
Although our controlled experiments have demonstrated that fMRI signals carry rich multi-object category semantics and coarse spatial location cues, we have not yet fully exploited these findings to optimize existing models and resolve the hierarchical disentanglement challenge posed at the outset. Future research will focus on accurately extracting such instance-level information and effectively incorporating it into visual signal decoding tasks.

\section{Conclusion}
In this paper, we establish an experimental framework to explore the brain’s visual perception mechanisms. By treating fMRI representations as an intervention condition and injecting them into multi-scale image features via cross attention, we compare downstream performance and intermediate feature changes on object detection and instance segmentation tasks with and without fMRI information. Our results demonstrate that fMRI signals indeed carry rich multi-object semantic cues and coarse spatial location information. However, existing decoding models only partially leverage these signals and have yet to fully integrate and apply them. This study offers new insights for future model optimization, advancing the efficient fusion and decoding of multi-granularity brain–vision information.

\bibliographystyle{splncs04_unsort}
\bibliography{reference}






\newpage
\appendix
\setcounter{section}{0}
\renewcommand\thesection{\Alph{section}}

\section{Downstream Loss}
\begin{equation}L_{\mathrm{det}}=\frac{1}{R_{\mathrm{els}}}\sum_{i=1}^{R}L_{\mathrm{els}}(p_{i},\hat{p_{i}})+\frac{\lambda}{R_{\mathrm{reg}}}\sum_{i=1}^{R}[p_{i}^{*}>0]L_{\mathrm{reg}}(t_{i},t_{i}^{*}).\end{equation}

- $p_{i}$:predicted class-score vector for proposal $i$;$p_{i}^{*}$: one-hot ground-truth.\\
- $L_{cls}$ is the cross-entropy loss.\\
- $t_i$ and $t_{i}^{*}$ are the predicted and ground-truth box-regression targets; $[p_{i}^{*}>0]$ is 1 for positive (foreground) proposals and 0 otherwise.\\
- $L_{reg}$ is the smooth-$L_1$ loss, and $\lambda$ weights its contribution.\\

\begin{equation}\mathcal{L}_{\mathrm{mask}}=\frac{1}{R_{\mathrm{mask}}}\sum_{i=1}^R[p_i^*>0]L_{\mathrm{BCE}}(m_i,m_i^*),\end{equation}

- $m_i\in{\mathcal{[0,1]}^{m\text{×}m}}$ is the predicted mask for proposal i; $m_i^*\in{\mathcal{[0,1]}^{m\text{×}m}}$ is the ground-truth binary mask.\\
- $L_{BCE}$ denotes the per-pixel binary cross-entropy loss.\\

Hyper-parameters $R_{cls}$ , $R_{reg}$ , and $R_{mask}$ normalize each term (usually the number of sampled RoIs), and $\lambda$ (often set to 1) balances classification vs. regression. This joint loss drives the model to correctly classify and localize objects while also learning precise instance masks.

\section{Ablation for LoRA’s Rank.}
A larger rank implies that the plugin introduces more parameters. We perform an ablation study on LoRA’s rank to determine its optimal value, searching over all experiments using fixed hyperparameters. As shown in Table ~\ref{tab:table2}, the best performance occurs at rank 16. However, we observe that all tested ranks yield comparable results, indicating that our method is largely insensitive to this hyperparameter. This aligns with previous LoRA research, which also found that increasing the rank does not significantly improve performance.
\begin{table}[]
\caption{Ablation experiments for rank. 
\label{tab:table3}}
\centering
\tabcolsep=0.4cm
\renewcommand\arraystretch{1.0}
\begin{tabular}{cccccccccccccc}
\hline
 &$r$                  & $det AP$   & $det AR_1$ & $seg AP$ & $seg AR_1$  \\ \hline
  & 2                                        & 20.5    & 21.0    & 18.5   & 19.3     \\
& 4                                          & 20.7   & 20.7    & 18.8    & 19.6      \\
 & 8                                 & \textbf{20.9}    & 20.9   & 18.7    & 19.7    \\ 
 & 16                                        & \textbf{20.9}   & \textbf{21.3}   & \textbf{18.8}    & \textbf{19.8}     \\
& 32                                        & 20.8    & 21.1    & 18.7    & 19.6     \\ \hline  
\end{tabular}
\end{table}

\section{Instance segmentation results.}
\begin{figure}[!t]
    \centering
    \includegraphics[width=0.7\linewidth]{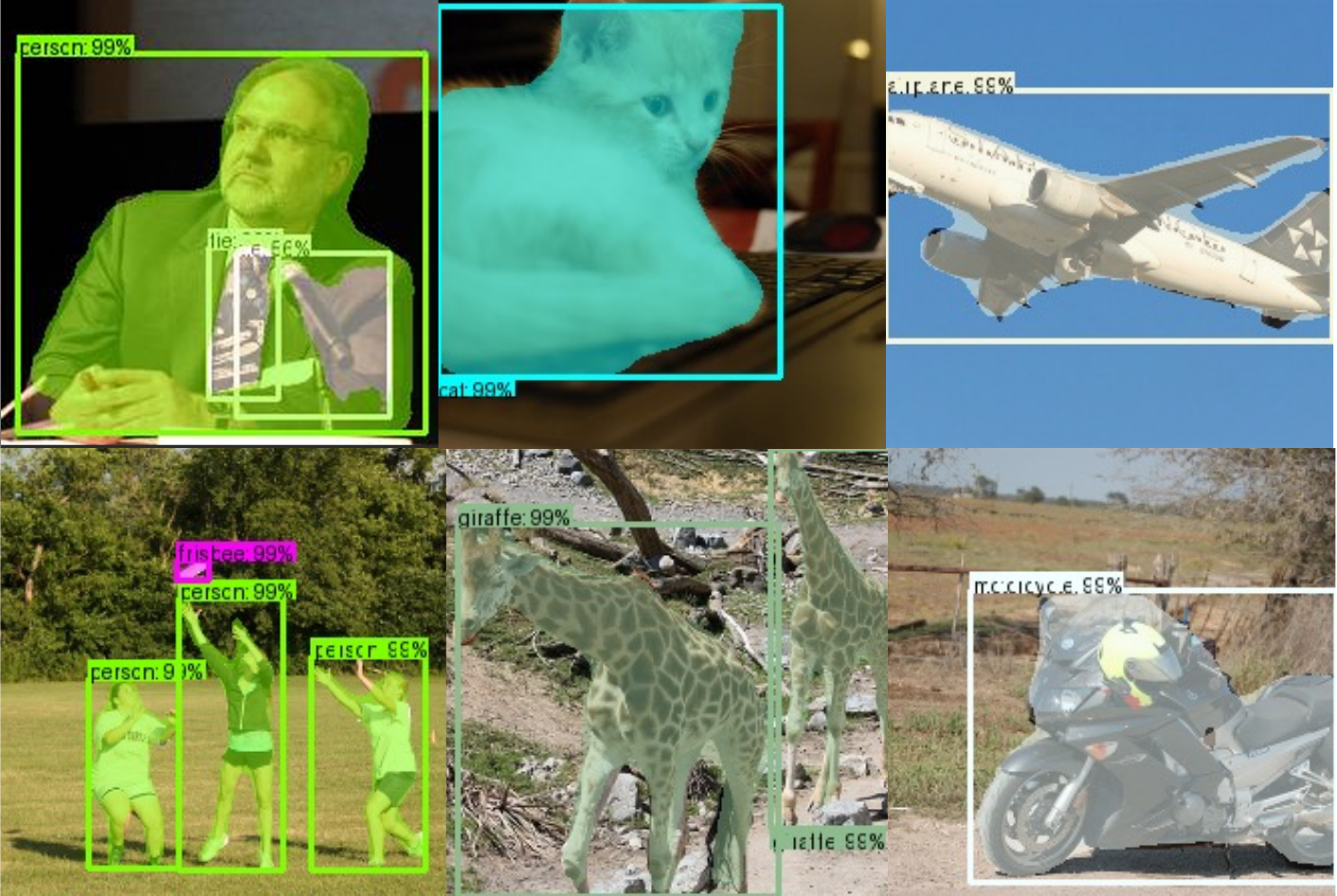}
    \caption{Instance segmentation result, including different types and different target quantities.}
    \label{fig9}
\end{figure}

In this section, we present the instance segmentation results of the model under various types of targets. It can be seen from Fig.~\ref{fig9} that all the targets in the pictures have been effectively recognized, and accurate masks have been generated

\begin{figure}[!b]
    \centering
    \includegraphics[width=1\linewidth]{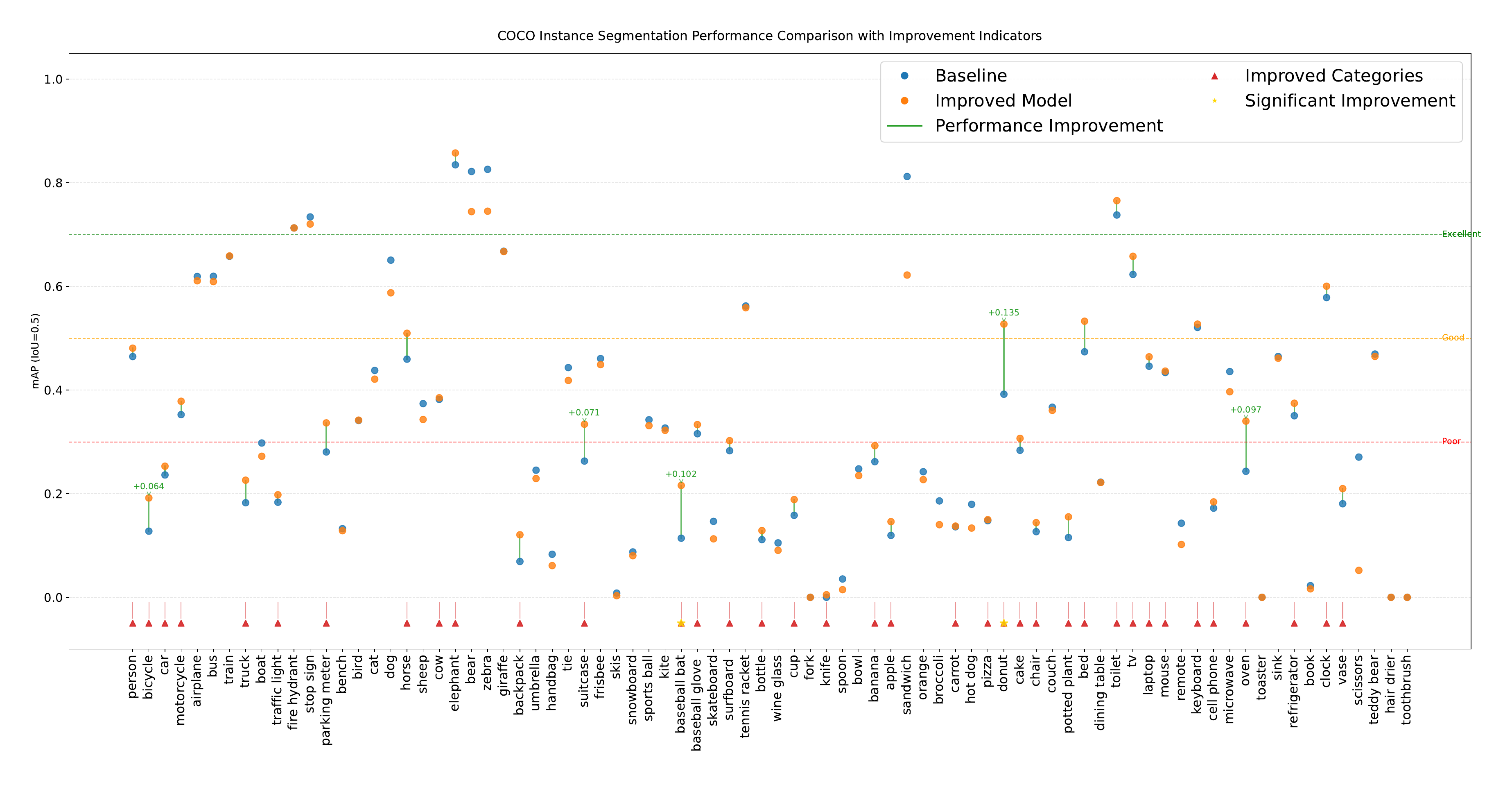}
    \caption{mAP point map presented by category (IoU$>$50). Our model has been improved in most categories.}
    \label{fig10}
\end{figure}
\section{Instance Segmentation Category Results.}
Although the mAP of most categories in the instance segmentation task has still been improved as shown in Fig.~\ref{fig10}, , the absolute value of the mAP is at a relatively low level compared to the instance segmentation task, which is in line with the fact that fMRI information contains rich category semantics as mentioned in our analysis, but it is lacking in fine-grained information representation.

\section{More Cross-Attention Heatmaps.}
Here, we adopt the average activation map of a single image. It can be seen that there are obvious differences in activation patterns among different types of images, and when there are multiple targets in the picture, more activations will be obtained.
\begin{figure}[!b]
    \centering
    \includegraphics[width=0.9\linewidth]{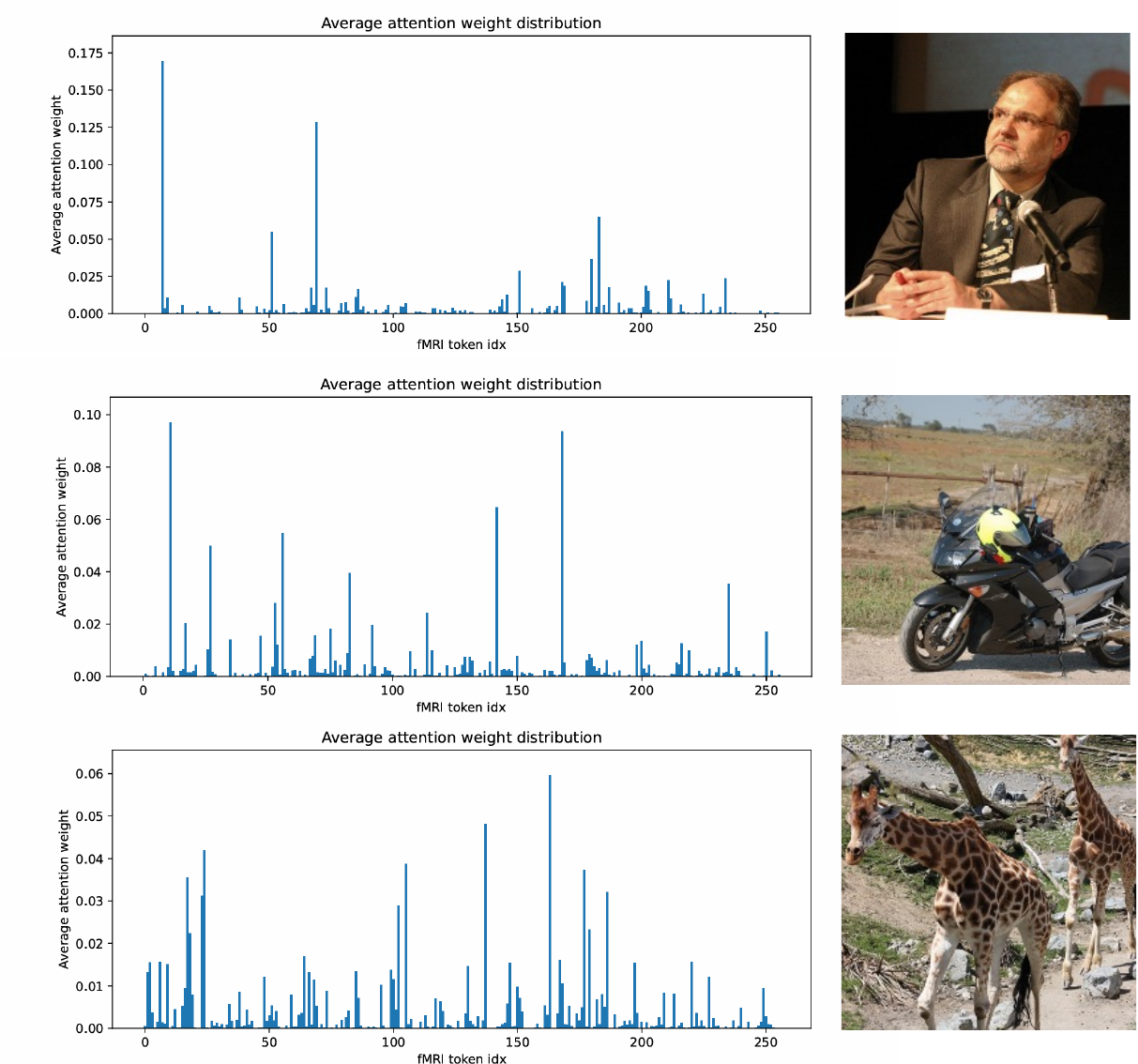}
    \caption{Multi-layer average attention weight maps. The selected categories from top to bottom are aircraft, giraffes, and motorcycles in sequence.}
    \label{fig12}
\end{figure}
\end{CJK}
\end{document}